\pdfoutput=1
\documentclass[11pt]{article}

\usepackage{acl}

\usepackage{times}
\usepackage{latexsym}
\usepackage{graphicx}
\usepackage{amsmath}
\usepackage{amsfonts}
\usepackage{algorithm}
\usepackage{algpseudocode}
\usepackage{booktabs}
\usepackage{multicol}
\usepackage{multirow}
\usepackage{enumitem}
\usepackage{xcolor,soul}
\usepackage{ulem}
\usepackage{tikz}
\usepackage{color, colortbl}

\usepackage{footnote}
\makesavenoteenv{tabular}
\makesavenoteenv{table}

\usepackage[T1]{fontenc}
\usepackage[utf8]{inputenc}

\usepackage{microtype}

\newcommand\Tstrut{\rule{0pt}{2.6ex}}       % "top" strut
\newcommand\Bstrut{\rule[-0.9ex]{0pt}{0pt}} % "bottom" strut
\newcommand{\TBstrut}{\Tstrut\Bstrut} % top&bottom struts

\newcommand{\tabitem}{~~\llap{\textbullet}~~}
\newcommand{\GS}{{Geodesic Summarizer}}
\newcommand{\X}{{GeoSumm}}
\newcommand{\SPACE}{\textsc{Space}}
\newcommand{\AMZN}{\textsc{Amazon}}
\newcommand{\OPO}{\textsc{OpoSum+}}

\usepackage{amssymb}
\usepackage{pifont}

\definecolor{darkgray2}{rgb}{0.36, 0.36, 0.36}
\definecolor{LightCyan}{rgb}{0.8,0.9,0.8}
\definecolor{LightRed}{rgb}{1,0.75,0.75}
\definecolor{teal}{rgb}{0.98, 0.75, 0}
\definecolor{Gray}{gray}{0.93}
\definecolor{mintbg}{rgb}{.63,.79,.95}

\newcommand{\light}[1]{\textcolor{darkgray2}{#1}}

\newcount\Comments  % 0 suppresses notes to selves in text
\definecolor{darkgreen}{rgb}{0,0.5,0}
\definecolor{lightgreen}{rgb}{0.63,0.89,0.72}
\definecolor{darkred}{rgb}{0.7,0,0}
\definecolor{tealtwo}{rgb}{0.1,0.6,0.7}
\definecolor{blue}{rgb}{0.0,0.1,0.9}
\definecolor{yellow}{rgb}{0.98,0.95,0.1}
\definecolor{orange}{rgb}{1.,0.7,0.0}
\definecolor{lightblue}{rgb}{0.70, 0.80, 0.89}
\definecolor{lightred}{RGB}{214,39,40}
\definecolor{darkblue2}{RGB}{31,119,180}
\definecolor{darkyellow}{RGB}{160,161,36}

\makeatletter
\def\@fnsymbol#1{\ensuremath{\ifcase#1\or \dagger\or \ddagger\or
\mathsection\or \mathparagraph\or \|\or **\or \dagger\dagger
\or \ddagger\ddagger \else\@ctrerr\fi}}
\makeatother

\title{Unsupervised Opinion Summarization Using Approximate Geodesics}

\author{Somnath Basu Roy Chowdhury\thanks{\hspace{0.1cm} Work done during an internship at Google Research.} \textsuperscript{,$1$}
\qquad Nicholas Monath$^{2}$ \qquad Avinava Dubey$^{2}$\\ 
\qquad \textbf{Amr Ahmed}$^{2}$ \qquad \textbf{Snigdha Chaturvedi}$^{1}$ 
\\ $^{1}$UNC Chapel Hill,  $^{2}$Google Research
\\ \texttt{\{somnath, snigdha\}@cs.unc.edu}
\\ \texttt{\{nmonath, avinavadubey, amra\}@google.com}
}

\begin{document}
\maketitle
\begin{abstract}
Opinion summarization is the task of creating summaries capturing popular opinions from user reviews.
In this paper, we introduce {\GS} ({\X}), a novel system to perform unsupervised extractive opinion summarization.  {\X} consists of an encoder-decoder based representation learning model that generates topical representations of texts. These representations capture the underlying semantics of the text as a distribution over learnable latent units. {\X} generates these topical representations by performing dictionary learning over pre-trained text representations at multiple layers of the decoder.
We then use these topical representations to quantify the importance of review sentences using a novel approximate geodesic distance-based scoring mechanism. We use the importance scores to identify popular opinions in order to compose general and aspect-specific summaries. Our proposed model, {\X}, achieves strong performance on three opinion summarization datasets. We perform additional experiments to analyze the functioning of our model and showcase the generalization ability of {\X} across different domains.
\end{abstract}

\section{Introduction}

As more and more human interaction takes place online, consumers find themselves wading through an ever-increasing number of documents (e.g., customer reviews) when trying to make informed purchasing decisions. As this body of information grows, so does the need for automatic systems that can summarize it in an unsupervised manner.
Opinion summarization is the task of automatically generating concise summaries from online user reviews~\cite{hu2004mining, pang2008lee, medhat2014sentiment}. For instance, opinion summaries allow a consumer to understand product reviews without reading all of them. 
Opinion summaries are also useful for sellers to receive feedback, and compare different products. The recent success of deep learning techniques has led to {a} significant improvement in summarization~\cite{rush2015neural, nallapati2016abstractive, cheng2016neural, see2017get, narayan2018ranking, liu2018generating} in supervised settings. However, it is difficult to leverage these techniques for opinion summarization due to the scarcity of annotated data. It is expensive to collect good-quality opinion summaries as human annotators need to read hundreds of reviews to write a single summary~\cite{moussa2018survey}. Therefore, most works on opinion summarization tackle the problem in an unsupervised {setting}.

Recent works~\cite{amasum, amplayo-etal-2021-aspect} focus on abstractive summarization, where fluent summaries are generated using novel phrases. However, these approaches suffer from issues like text hallucination~\cite{rohrbach-etal-2018-object} that affect the faithfulness of generated summaries~\cite{faithfulness}.  Extractive summaries are less prone to these problems presenting the user with a representative subset of the original reviews.

We focus on the task of unsupervised extractive opinion summarization, where the system selects sentences representative of the user opinions.  Inspired by previous works~\cite{semae, qt}, we propose a novel encoder-decoder architecture along with objectives for (1) learning sentence representations that capture the underlying semantics, and (2) a sentence selection algorithm to compose a summary.

One of the challenges in extractive summarization is quantifying the importance of opinions. An opinion is considered to be important if it is semantically similar to opinions from other users. Using off-the-shelf pre-trained representations to obtain semantic similarity scores has known issues~\cite{bark}. These similarity scores can behave counterintuitively due to the high anisotropy of the representation space (a few dimensions dominate the cosine similarity scores). 
Therefore, we {use} topical representations~\cite{lda}, which capture the  semantics of text as a distribution over latent semantic units. These semantic units encode underlying concepts or topics. The semantic units can be captured using a learnable dictionary~\cite{Engan1999MethodOO,mairal2009online,aharon2006k, lee2006efficient}. Topical representations enable us to effectively measure semantic similarity between text representations as they are distributions over the same support. Text representations from reviews lie on a high-dimensional manifold. It is important to consider the underlying manifold while computing the importance score of a review. Therefore, we use the approximate geodesic distance between topical text representations to quantify the importance scores of reviews.

In this paper, we present {\textbf{Geo}desic \textbf{Summ}arizer} ({\X}) that learns topical text representations in an unsupervised manner from distributed representations~\cite{hinton1984distributed}. 
We also present a novel sentence selection scheme that compares topical sentence representations in high-dimensions using approximate geodesics. Empirical evaluations show that {\X} achieves strong performance on three opinion summarization datasets -- {\OPO}~\cite{amplayo-etal-2021-aspect}, {\AMZN}~\cite{he2016ups} and {\SPACE}~\cite{angelidis2021extractive}. 
Our primary contributions are:

\begin{itemize}[topsep=1pt, leftmargin=*, noitemsep]
    \itemsep0mm
    \item We present an extractive opinion summarization system, {\X}. It consists of an unsupervised representation learning system and a sentence selection algorithm  (Section~\ref{sec:gs}). 
    \item We present a novel representation learning model that learns topical text representations from distributed representations using dictionary learning  (Section~\ref{subsec:unsup}). 
    \item We present a novel sentence selection algorithm that computes the importance of text using approximate geodesic distance  (Section~\ref{subsec:ss}). 
    \item {\X} achieves strong performance on 3 opinion summarization datasets (Section~\ref{sec:results}).
\end{itemize}

\section{Task Setup}
In extractive opinion summarization, the objective is to select representative sentences from a review set. Specifically, each dataset consists of a set of entities $\mathbf{E}$ and their corresponding review set $\mathcal{R}$.
For each entity $e \in \mathbf{E}$ (e.g., a {particular} hotel such as {the} Holiday Inn {in Redwood City, CA}), a review set $\mathcal{R}_e = \{r_1, r_2, \ldots\}$ is provided, where each review is an {ordered set} of sentences $r_i = \{s_1^{(i)}, s_2^{(i)}, \ldots\}$. For simplicity of notation, we will represent the set of review sentences corresponding to an entity $e$ as $\mathcal{S}_e=\bigcup_{r_i \in \mathcal{R}_e} r_i$. For each entity, reviews encompass a set of aspects $\mathcal{A}_e = \{a_1, a_2, \ldots\}$ (e.g.,
service, food of a hotel). In this work, we consider two forms of extractive summarization: (a) \textit{general summarization}, where the system selects a subset of sentences $\mathcal{O}_e \subset \mathcal{S}_e$, that best represents popular opinions in the review set $\mathcal{R}_e$; (b) \textit{aspect summarization}, where the system selects a representative sentence subset $\mathcal{O}_e^{(a)} \subset \mathcal{S}_e$,  about a specific aspect $a$ (e.g., service) of an entity $e$ (e.g., hotel).

\section{{\GS} ({\X})}
\label{sec:gs}

In this section, we {present our proposed approach}  {\GS} ({\X}). {\X} has two parts: (a) {an} unsupervised model to learn topical representations of review sentences, and (b) a sentence selection algorithm that uses the approximate geodesic distance between topical representations,  to compose the extractive summary.

\subsection{Unsupervised Representation Learning}
\label{subsec:unsup}
The goal of the representation learning model is to learn topical representations of review sentences. 
Topical representations model text as a distribution over underlying concepts or topics. This is useful for unsupervised extractive summarization because we want to capture the aggregate semantic distribution and quantify the importance of individual review sentences with respect to the aggregate distribution. Topical representations allow us to achieve both. Being a distribution over latent units, topical representations can be combined to form an aggregate (mean) representation, enabling compositionality. Also, it is convenient to measure the similarity between representations using conventional metrics (like cosine similarity).

\begin{figure}[t!]
    \centering
    \includegraphics[width=0.48\textwidth, keepaspectratio]{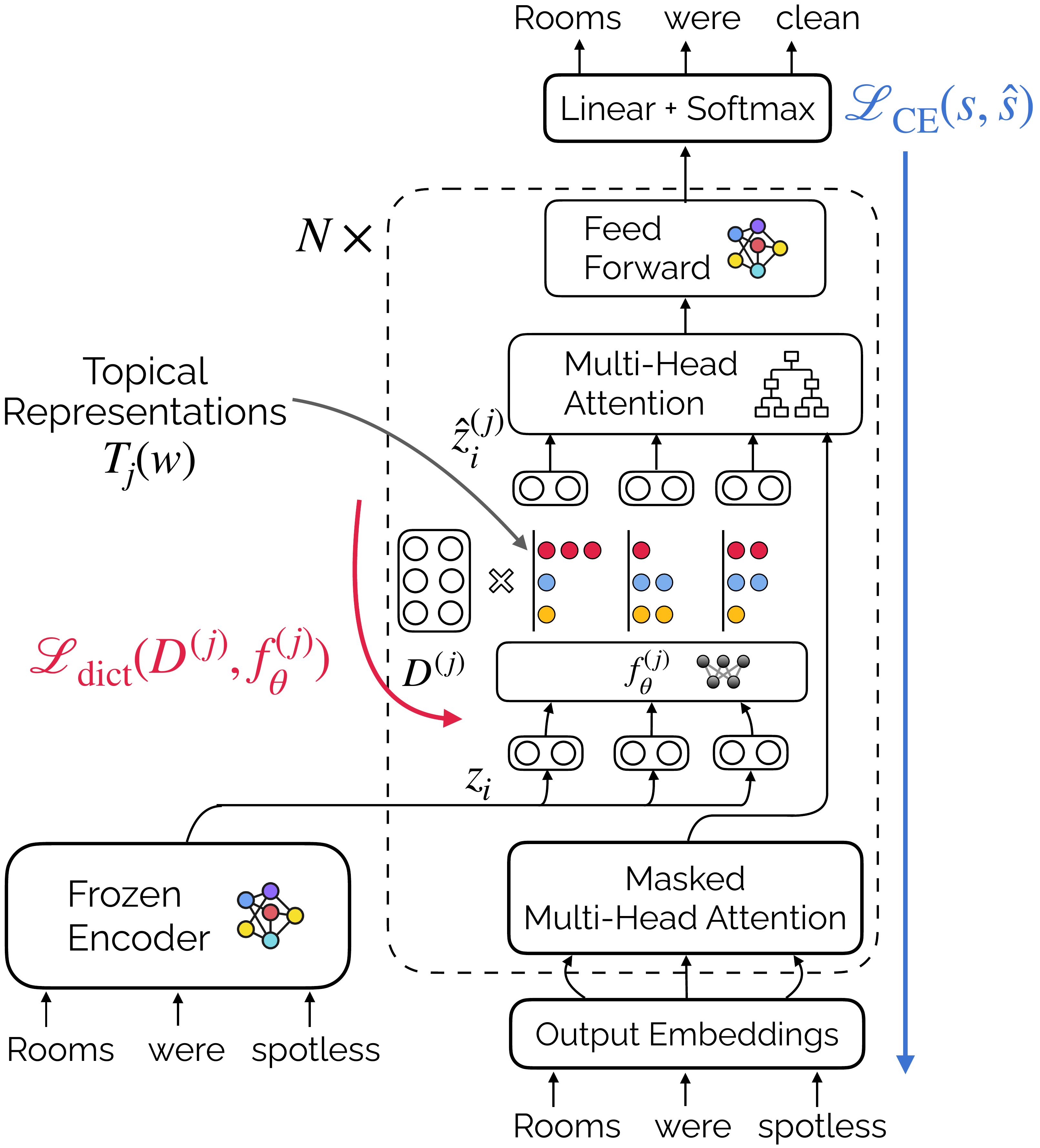}
    \caption{
    Architecture of {\GS}. Sparse representations of words are formed via the kernel function $f_\theta^{(j)}$. 
    The representations are trained to reconstruct the output embeddings of the encoder layer. 
    Alongside the dictionary learning objective, we use an unsupervised sentence reconstruction cross-entropy loss. 
    $N$ indicates the number of decoder layers.
    }
    \label{fig:model}
    \vspace{-12pt}
\end{figure}

We propose to model topical representations by decomposing pre-trained representations using dictionary learning~\cite{6873279, lotfi2018fast}. In this setup, the various components of the dictionary capture latent semantic units and we consider the representation 
over dictionary {elements} as the topical representation. Unlike conventional dictionary learning algorithms, we use a sentence reconstruction objective for learning the dictionary. We use an encoder-decoder architecture to achieve this. 
We retrieve word embeddings from a pre-trained encoder.
We modify the architecture of a standard Transformer decoder by introducing a dictionary learning component at each decoder layer. The pre-trained word embeddings {obtained from the encoder} are decomposed using these dictionary learning components to obtain topical representations. Then, we combine the topical word representations at different decoder layers to form a sentence representation.
The schematic diagram of the model is shown in Figure~\ref{fig:model}. Next, we will discuss each of the components in detail.

\noindent \textbf{Encoder}. We obtain contextual word embeddings from a pre-trained BART~\cite{bart} encoder. We keep the weights of the encoder frozen during training. In Section~\ref{sec:frozen}, we discuss why frozen representations are important for our model. 
Given an input sentence $s = \{w_1, \ldots, w_L\}$, we retrieve contextual word embeddings $z_i$'s from the BART encoder:

\begin{equation}
    {{z}_i} = \mathrm{sg}(\mathrm{enc}({\mathrm{w}_i})) \in \mathbb{R}^{d}
\end{equation}

\noindent where $\mathrm{sg}(\cdot)$ denotes the stop gradient operator.

\noindent \textbf{Dictionary Learning}. We describe the dictionary learning component within each decoder layer.  We use dictionary learning to decompose pre-trained word representations {from the encoder} to obtain a sparse representation for each word. We want word representations to be sparse because each word can capture only a small number of semantics. 
We forward word representations from the encoder to the decoder layers. 
For the $j$-th decoder layer, we use a dictionary, $\mathbf{D}^{(j)} \in \mathbb{R}^{m \times d}$, and kernel function, $k_j(\cdot, \cdot)$, where $j \in \{1, \ldots, N\}$ ($N$ is the number of decoder layers). 
The dictionary captures the underlying semantics in the text by enabling us to model text representations as a combination of dictionary elements. Specifically, we learn a topical word representation $T_j(w_i)$ over the dictionary $\mathbf{D}^{(j)}$ as: % shown below:

\begin{equation}
    \begin{aligned}
     &\hat{z_i}^{(j)} = {\mathbf{D}^{(j)}}^T T_j(w_i)\\
     &T_j(w_i) = k_j({z}_i, \mathbf{D}^{(j)}) \in \mathbb{R}^m,
    \end{aligned}
    \label{eqn:reconstruction}
\end{equation}

\noindent where $\hat{z}_i^{(j)}$  is the reconstructed word embedding, 
and $k_j(\cdot, \cdot) \in \mathbb{R}^m$ is the kernel function that measures the similarity between $z_i$ and individual dictionary elements. In practice, since the dictionary is common for all word embeddings $z_i$'s, the kernel function can be implemented as:

\begin{equation}
    k_j(\mathrm{z}_i, \mathbf{D}^{(j)}) = f_{\theta}^{(j)}(\mathrm{z}_i) \in \mathbb{R}^m,
\end{equation}
\noindent where $f_{\theta}^{(j)}$ is a feed-forward neural network with ReLU non-linearity. ReLU non-linearity ensures that the kernel coefficients are positive and also encourages sparsity.

Following conventional dictionary learning algorithms~\cite{fista}, the dictionary $\mathbf{D}^{(j)}$ and kernel layer $f_{\theta}^{(j)}$ are updated iteratively. We ensure the sparsity of the word representations $f_{\theta}^{(j)}(z)$ by adding an L1-penalty to the loss. Overall, this can be achieved by using the loss function:

\begin{equation*}
\begin{aligned}
    \mathcal{L}_{\mathrm{dict}}(\mathbf{D}^{(j)}, f_\theta^{(j)}) &= \| \mathrm{z}_i - \mathrm{sg}({\mathbf{D}^{(j)}}^T)f_{\theta}^{(j)}(\mathrm{z}_i) \|_2 \\ 
    & + \| \mathrm{z}_i - {\mathbf{D}^{(j)}}^T \mathrm{sg}(f_{\theta}^{(j)}(\mathrm{z}_i)) \|_2 \\
    & + \big\lvert f_{\theta}^{(j)}(z_i) - \mathbb{E}\left[f_{\theta}^{(j)}(z_i)\right] \big\rvert_1,
\end{aligned}
\end{equation*}

\noindent where the gradient update of the dictionary $\mathbf{D}^{(j)}$ and kernel layer $f_{\theta}^{(j)}$ are performed independently. 

\noindent \textbf{Decoder}. We build on the decoder architecture introduced by \citet{vaswani2017attention}. A decoder layer consists of 3 sub-layers (a) masked multi-head attention layer that takes as input decoder token embeddings, (b) multi-head attention that performs cross-attention between decoder tokens and encoder stack output, and (c) feed-forward network. We modify the cross attention multi-head sub-layer to attend over the reconstructed word embeddings $\hat{z}_i^{(j)}$ (Equation~\ref{eqn:reconstruction}), instead of the encoder stack output (shown in Figure~\ref{fig:model}). Finally, the decoder autoregressively generates the reconstructed sentence $\hat{s} = \{\hat{w_1}, \ldots, \hat{w_L}\}$.

\noindent \textbf{Training}. The system 
is trained using the sentence reconstruction objective. The overall objective function is shown below:

\begin{equation}
    \mathcal{L}_{\mathrm{CE}}(s, \hat{s}) + \sum\limits_{j=1}^{N} \mathcal{L}_{\mathrm{dict}}(\mathbf{D}^{(j)}, f_\theta^{(j)}),
\end{equation}

\noindent where $\mathcal{L}_{\mathrm{CE}}$ is the cross-entropy loss, and $f_\theta^{(j)}$ is the implementation of the  kernel function $k_j(\cdot, \cdot)$ corresponding to the $j$-th decoder layer. The above loss function is used to update the decoder, the dictionary elements, and the kernel parameters while keeping the encoder weights frozen.

\noindent \textbf{Sentence Representations}.\label{sec:sent-rep}
We combine topical word representations from different decoder layers to form a sentence representation.
First, we obtain a word representation, $T_j(w) \in \mathbb{R}^{m}$ from each decoder layer. 
We compose the final word representation $\mathbf{x}_w$ by concatenating representations from all decoder layers.

\begin{equation}
    \mathbf{x}_w = [T_1(w), \ldots, T_N(w)] \in \mathbb{R}^{mN},
\end{equation}

\noindent where $m$ is the dictionary dimension and $N$ is the number of decoder layers. We use max-pooling over the dimensions of word representations to form a sentence representation $\mathbf{x}_s$ as shown below.

\begin{equation}
\begin{aligned}
    &\mathbf{x}^s_{n} = \max_{w \in \{w_1, \ldots, w_L\}} \mathbf{x}_w\big\rvert_n \\
    &\bar{\mathbf{x}}_s = \{\mathbf{x}^s_{n}\}_{n=1}^{mN}, \mathbf{x}_s = {\bar{\mathbf{x}}_s}/{\| \bar{\mathbf{x}}_s \|_1} \in \mathbb{R}^{mN}, \\
\end{aligned}
\end{equation}

\noindent where $\mathbf{x}_w\big\rvert_n$ is the $n$-th entry of the vector $\mathbf{x}_w$. The sentence representation $\mathbf{x}_s$ is normalized to a unit vector. Next, we discuss how we leverage these topical sentence representations to compute importance scores using approximate geodesics. We use the importance scores to compose the final extractive summary for a given entity.

\begin{figure}[t!]
    \centering
    \includegraphics[width=0.43\textwidth, keepaspectratio]{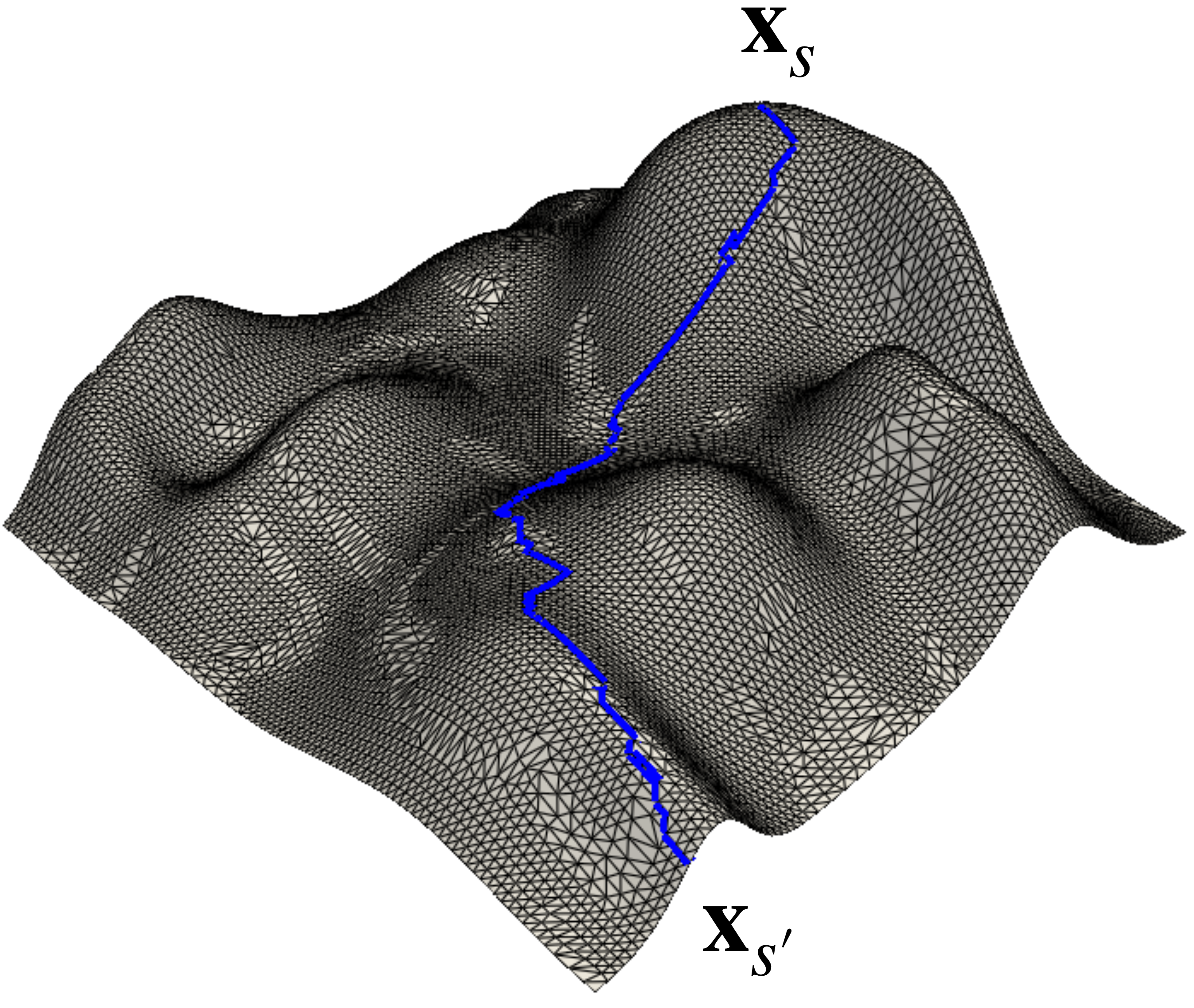}
    \caption{Illustration of the geodesic shortest path (shown in \textcolor{blue}{\textbf{blue}}) between two sentence representations $\mathbf{x}_s$ and $\mathbf{x}_{s'}$ on a three-dimensional manifold.}
    \label{fig:geodesic}
\end{figure}

\subsection{General Summarization}
\label{subsec:ss}

We use representations retrieved from {\X} to select sentences representative of popular opinions in the review set. For an entity $e$, the set of sentence representations is denoted as $\mathcal{X}_e = \{\mathbf{x}_s | s \in \mathcal{S}_e\}$. For a summary budget $q$, we select a subset of sentences $\mathcal{O}_e \subset \mathcal{S}_e$ according to their importance scores, such that $|\mathcal{O}_e| = q$. First, we compute a mean representation as shown: ${\mu_e} = \mathbb{E}_{s \sim \mathcal{S}_e}[\mathbf{x}_s]$.
Secondly, we define the importance of a sentence $s$, as the distance from the mean representation $d(\mathbf{x}_s, \mu_e)$. However, we do not directly evaluate $d(\cdot, \cdot)$ using a similarity metric. Representations in $\mathcal{X}_e$ lie in a high-dimensional manifold,
and we aim to measure the geodesic distance \cite{geodesic} between two points along that manifold. An illustration of the geodesic distance between two points is shown in Figure~\ref{fig:geodesic}. Computing the exact geodesic distance is difficult 
without explicit knowledge of the manifold structure~\cite{surazhsky2005fast}. 
We approximate the manifold structure using a $k$-NN graph. Each sentence representation forms a node in this graph.  A directed edge exists between two nodes if the target node is among the $k$-nearest neighbours of the source node. The edge weight between two nodes $(s, s')$  is defined using their cosine similarity distance, $d(s, s') = 1 - \mathbf{x}_s\mathbf{x}_{s'}^T$.   
The geodesic {distance} between two sentence representations is computed using the shortest path distance along the weighted graph. Therefore, the importance score $I(s)$ for a sentence $s$, is defined as:
\begin{equation}
    {I}(s) = 1/\mathrm{ShortestPath}(\mathbf{x}_s, \mu_e),
    \label{eqn:rel}
\end{equation}
\noindent where the shortest path distance is computed using Dijkstra's algorithm \cite{dijkstra1959note}. We select the top-$q$ sentences according to their importance scores $I(s)$ {to form the final general extractive summary}. The overall sentence selection routine is shown in Algorithm~\ref{alg:summ}. 

\begin{algorithm}[t!]
\caption{General Summarization Routine}
\label{alg:unconstrained}
\begin{algorithmic}[1]
    \State \textbf{Input}: A set of sentence representations $\mathcal{X}_e = \{\mathbf{x}_s| s \in \mathcal{S}_e\}$ are review sentences for entity $e$.
    \State $\mu_e \leftarrow \mathbb{E}_{s \sim \mathcal{S}_e}[\mathbf{x}_s]$
    \State $\mathbf{A} \leftarrow \mathrm{knn}(\mathcal{X}_e \cup \mu_e) \in \mathbb{R}^{l \times l}$ \Comment{adjacency matrix of $k$-NN graph, $l = \lvert S_e \rvert + 1$.}
    \State $d \leftarrow  \mathrm{Dijkstra}(\mathbf{A}, \mu_e)$ \Comment{shortest distances of all nodes from $\mu_e$}
    \State $I \leftarrow \{1/d(s) | s \in \mathcal{S}_e\}$ \Comment{importance scores}
    \State ${t}_q \leftarrow  \min \text{top-}q(I)$ \Comment{top-$q$ threshold}
    \State $\mathcal{O}_e \leftarrow \{s \;|\; I(s) \geq t_q,  s \in \mathcal{S}_e\}$
    \State \Return $\mathcal{O}_e$
\end{algorithmic}
\label{alg:summ}
\end{algorithm}

\subsection{Aspect Summarization}
In aspect summarization, the goal is to select representative sentences to form a summary specific to an aspect (e.g., durability) of an entity (e.g., bag).
To perform aspect summarization, we compute the mean representation of aspect-specific sentences as shown:
 $\mu_e^{(a)} = \mathbb{E}_{s \sim \mathcal{S}_e^{(a)}}[\mathbf{x}_s]$, where $\mathcal{S}_e^{(a)}$ is the set of sentences mentioning aspect $a$. We identify $\mathcal{S}_e^{(a)}$ by detecting the presence of aspect-specific keywords available with the dataset. To ensure the selected sentences are aspect-specific,
 we introduce a measure of \textit{informativeness} \cite{semae, peyrard-2019-simple}. 
 Informativeness penalizes a sentence for being close to the overall mean $\mu_e$.
 Therefore, we model the aspect-specific importance score $I_a(s)$ as:
 \vspace{-15pt}

\begin{equation}
    I_{a}(s) = 1/\mathrm{ShortestPath}(\mathbf{x}_s, \mu_e^{(a)}) - \gamma I(s),
    \label{eqn:gamma}
\end{equation}

\noindent where $\gamma$ is a hyperparameter, $I(s)$ is the overall importance score (obtained from Eqn.~\ref{eqn:rel}). Aspect summary $\mathcal{O}_e^{(a)}$ is composed using the top-$q$ sentences according to the aspect-specific scores, $I_a(s)$. 

\section{Experiments}
We evaluate the performance of {\X} on extractive summarization. Given a set of user reviews the system needs to select a subset of the sentences as the summary. This summary is then compared with human-written summaries.
In this section, we discuss the experimental setup in detail.

\subsection{Datasets \& Metrics}
We evaluate {\X} on three publicly available opinion summarization datasets: 

\noindent (a) {\OPO}~\cite{amplayo2021aspect} is an extended 
version of the original \textsc{OpoSum} dataset~\cite{oposum}. This dataset contains Amazon reviews from six product categories (like laptops, bags, etc.), with 3 human-written summaries in the test set. The extended version contains additional product reviews and aspect-specific human annotations.

\noindent (b) {\AMZN}~\cite{he2016ups, brazinskas} has product reviews of 4 different categories (like electronics, clothing, etc.) from Amazon, with 3 human summaries per entity.

\noindent(c) {\SPACE}~\cite{qt} contains reviews for hotels from Tripadvisor. {\SPACE} provides three human-written abstractive summaries and six aspect-specific summaries per hotel entity.

Statistics of the datasets are provided in Table~\ref{tab:stat}. We observe that {\SPACE} dataset has significantly more reviews per entity compared to other datasets.

\begin{table}[t!]
    \small
    \centering
	\resizebox{0.48\textwidth}{!}{
		\begin{tabular}{ l c c c} 
			\toprule[1pt]
			Dataset & {Reviews} & {Train / Test Ent.} & {Rev./Ent.} \\ 			
			\midrule[1pt]
			{\OPO} & 4.13M & 95K / 60 & 10\\
			{\AMZN} & 4.75M & 183K / 60 & ~8\\
			{\SPACE} &  1.14M & 11.4K / 50 & 100\\
			\bottomrule[1pt]
	\end{tabular}
}
	\vspace{-5pt}
\caption{ Dataset statistics for {\OPO}, {\AMZN} and {\SPACE} datasets. (Train/Test Ent.: Number of entities in the \textit{training} and \textit{test} set; Rev./Ent.: Number of reviews per entity in the \textit{test} set.)}
\label{tab:stat}
\vspace{-10pt}
\end{table}

\begin{table*}[t!]
    \centering
    \resizebox{0.99\textwidth}{!}{
		\begin{tabular}{ @{}cl@{} c c c | c c c | c c c } 
			\toprule[1pt]
            & \multirow{2}{*}{Method} & \multicolumn{3}{c|}{{\OPO}} & \multicolumn{3}{c|}{{\AMZN}} & \multicolumn{3}{c}{{\SPACE}}\\
			& & R1 & R2 & RL & R1 & R2 & RL & R1 & R2 & RL \\ 			
			\midrule[1pt]
		    \parbox[t]{0.6mm}{\multirow{3}{*}{\rotatebox[origin=c]{90}{\small{Single Rev.}}}} 
		    & Random & 29.88 & 5.64 & 17.19 & 27.66 & 4.72 & 16.95  & 26.24 & 3.58 & 14.72 \Tstrut\\
		    & Centroid\textsubscript{BERT} & 33.44 & 11.00 & 20.54 & 29.94 & 5.19 & 17.70 & 31.29 & 4.91 & 16.43\\
		    & Oracle & \light{32.89} & \light{23.20} & \light{28.73} & \light{31.69} & \light{6.47} & \light{19.25} &  \light{33.21} & \light{8.33} & \light{18.02} \Bstrut\\
		    \midrule[1pt]
		    \parbox[t]{0.6mm}{\multirow{7}{*}{\rotatebox[origin=c]{90}{\small{Abstractive}}}} 
		    & Opinosis \cite{ganesan2010opinosis} & - & - & - & 28.42 & 4.57 & 15.50 & 28.76 & 4.57 & 15.96  \Tstrut\\
    		& MeanSum \cite{chu2019meansum} & 34.95 & 7.49 & 19.92 & 29.20 & 4.70 & 18.15 & 34.95 & 7.49 & 19.92\\
    		& {Copycat \cite{bravzinskas2019unsupervised} } & \underline{36.66} & 8.87 & \underline{20.90} &  {31.97} & {5.81} & {20.16}  & 36.66 & 8.87 & 20.90\\ 
    		& {PlanSum \cite{amplayo2021unsupervised}} & - & - & - & 32.87 & 6.12 &  19.05 & - & - & - \\
    		& {TranSum \cite{wang-wan-2021-transsum} } & - & - & - & 34.23 & \underline{7.24} & 20.49 & - & - & -\\
		& {\textsc{Coop} \cite{iso-etal-2021-convex-aggregation} } & - & - & - &  \underline{36.57} & {7.23} & \underline{21.24}& - & - & - \\
    		& {AceSum \cite{amplayo2021unsupervised} } & 32.98 & \underline{10.72} & 20.27 & - & - & - & \underline{40.37} & \underline{11.51} & \underline{23.23}\Bstrut\\
		    \midrule[1pt]
		    \parbox[t]{0.6mm}{\multirow{4}{*}{\rotatebox[origin=c]{90}{\small{Extractive}}}} 
	    	& LexRank\textsubscript{BERT}~\cite{erkan2004lexrank}\hspace*{0.15cm} & 35.42 & 10.22 & 20.92 & 31.47 & 5.07 & {16.81}  & 31.41 & ~~5.05 & 18.12 \Tstrut\\
			& QT~\cite{qt} & 37.72 & 14.65 & 21.69 & {31.27} & 5.03 & 16.42 &  38.66 & 10.22 & 21.90\\
			& AceSum\textsubscript{EXT}~\cite{amplayo2021aspect} & 38.48 & 15.17 & 22.82 & -  & - & - & 35.50 & ~~7.82 & 20.09\\
			& SemAE~\cite{semae} & 39.16 & 16.85 & 23.61 & {32.03} & 5.38 & 16.47 & \textbf{42.48} & \textbf{13.48} & \textbf{26.40} \Bstrut\\
			\midrule[1pt]
			& {\GS} (\X) & \textbf{41.55}  & \textbf{20.77} & \textbf{25.19} & \textbf{33.75}  & \textbf{7.15} &  \textbf{18.79} & {42.36} & 12.44 & {24.80} \TBstrut\\
			\bottomrule[1pt]
	\end{tabular}
}
    \vspace{-5pt}
    \caption{Evaluation results of {\X} and baseline approaches on general summarization. We observe that {\X} achieves strong performance on all datasets. 
    We report the ROUGE-F scores denoted as -- R1: ROUGE-1, R2: ROUGE-2, RL: ROUGE-L. We highlight the best performance achieved by an extractive summarization system in \textbf{bold} and the best abstractive summarization performance in \underline{underline}.}
    \label{tab:results}
    \vspace{-10pt}
\end{table*}

\subsection{Implementation Details}
Our experiments are implemented using the TensorFlow~\cite{tf} framework. We use BART\textsubscript{base}~\cite{bart} architecture as our encoder-decoder model. We initialize the encoder with pre-trained weights from BART, while the decoder is trained from scratch. In our experiments, we use dictionary dimension $m=8192$, number of decoder layers $N=6$, and hidden dimension $d=768$. {\X} was trained for 15K steps on 16 TPUs in all setups. We optimize our model using Adam~\cite{adam} optimizer with a learning rate of $10^{-5}$. We set aspect-summarization parameter $\gamma=0.5$ for {\OPO} and $\gamma=0.7$ for {\SPACE} (Equation~\ref{eqn:gamma}). All hyperparameters were tuned using grid-search on the development set. We will make our code publicly available.

\subsection{Baselines}
We compare GeoSumm with several summarization  systems (including the current state-of-the-art)  that  can  be  classified  into three broad categories:

\begin{table}[t!]
    \centering
	\resizebox{0.49\textwidth}{!}{
    \begin{tabular}{ @{}cl@{} c c c  c c c} 
        \toprule[1pt]
        & \multirow{2}{*}{Method} & \multicolumn{3}{c}{\OPO} & \multicolumn{3}{c}{\SPACE} \\
        & & R1 & R2 & RL & R1 & R2 & RL\\
        \midrule[1pt]
        \parbox[t]{0.1mm}{\multirow{3}{*}{\rotatebox[origin=c]{90}{\small{Abstract.}}}} 
        & MeanSum  & 24.63 & 3.47 & 17.53 & 23.24 & 3.72 & 17.02\\
        & CopyCat  &  26.17 & 4.30 & 18.20 & 24.95 & 4.82 & 17.53\\
        & AceSum & 29.53 & 6.79 & 21.06 & \textbf{32.41} & 9.47 & \textbf{25.46}\\
        \midrule[0.5pt]
        \parbox[t]{0.1mm}{\multirow{4}{*}{\rotatebox[origin=c]{90}{\small{Extractive}}}} 
        & LexRank &  22.51 & 3.35 &  17.27 & 27.72 & 7.54 & 20.82\\
        & QT & 23.99 & 4.36 & 16.61 & 28.95 & 8.34 & 21.77\\
        & SemAE & 25.30 & 5.08 & 17.62 & 31.24 & \textbf{10.43} & 24.14\\
        & AceSum\textsubscript{EXT}\hspace*{0.1cm} & 26.16 & 5.75 & 18.55 & 30.91 & 8.77 & 23.61\\
        \midrule[0.5pt]
        & {\X} & \textbf{30.84} & \textbf{8.98} & \textbf{21.64} & 26.61 & 5.82 & 19.37\\
        \bottomrule[1pt]
\end{tabular}
}

	\vspace{-5pt}
\caption{ Evaluation results on aspect summarization. The best scores for each metric is highlighted in \textbf{bold}. {\X} achieves the state-of-the-art performance on {\OPO}, while achieving competitive performance with other extractive methods on {\SPACE}.
}
\label{tab:asp-results}
\vspace{-10pt}
\end{table}

\noindent $\bullet$ \textit{Single Review} systems select a single review as the summary. We compare with the following systems: (a) \textit{Random} samples a review randomly from the review set; (b) \textit{Centroid} selects a review closest to the centroid of the review set. The centroid is computed using BERT~\cite{bert} embeddings; (c) \textit{Oracle} selects the best review based on ROUGE overlap with the human-written summary. 

\noindent $\bullet$ \textit{Abstractive} systems generate summaries using novel phrasing. We compare {\X} with the following systems: MeanSum \cite{chu2019meansum}, {Copycat} \cite{bravzinskas2019unsupervised}, PlanSum~\cite{amplayo2021unsupervised}, TranSum~\cite{wang-wan-2021-transsum}, {\textsc{Coop} \cite{iso-etal-2021-convex-aggregation}}, and {AceSum} \cite{amplayo2021aspect}.

\noindent $\bullet$ \textit{Extractive} systems select text phrases from the review set to form the summary. We compare with the following systems: LexRank \cite{erkan2004lexrank} using BERT embeddings, QT \cite{qt}, AceSum\textsubscript{EXT} \cite{amplayo2021aspect}, and SemAE \cite{semae}.

\subsection{Results}
\label{sec:results}
We discuss the performance of {\X} on general and aspect-specific summarization. 
We evaluate the quality of the extracted summaries using the automatic metric -- ROUGE F-scores~\cite{rouge}, which measures the n-gram overlap with the human-written summaries. 

\begin{table}[t!]
    % \small
	\centering
    \resizebox{0.38\textwidth}{!}{
	\begin{tabular}{ l c  c c} 
		\toprule[1pt]
		 General  & {Inform.} & {Coherence} & {Redund.} \\ 			
		\midrule[1pt]
        {SemAE}  & -7.3 & -10.0 & -51.4\\
		{QT} &  ~\textbf{8.0} & ~-4.7 & {12.7}\\
		{\X} &  {-0.7} & ~~\textbf{14.7}* & ~~\textbf{38.7}*\Bstrut\\
		\bottomrule[0.5pt]
	\end{tabular}
}

    \vspace{-5pt}
	\caption{Human evaluation results of general summarization for \textsc{Space} dataset. (*): statistically significant difference with all baselines ($p < 0.05$, using paired bootstrap resampling \citet{koehn2004statistical}).}
	\label{tab:human-eval-general}
	\vspace{-10pt}
\end{table}

\noindent \textbf{General Summarization}. We present the results of {\X} and baseline approaches on general summarization in Table~\ref{tab:results}. We observe that {\X} achieves strong performance across all datasets. For {\OPO} and {\AMZN} datasets, {\X} achieves significant improvement over baselines achieving the best performance among extractive summarization systems.  
For the {\SPACE} dataset, it is competitive with baselines falling slightly short of the state-of-the-art model, SemAE. However, we observe that {\X}'s summaries are much more diverse leading to significantly better human evaluation scores compared to SemAE.

\noindent \textbf{Aspect Summarization}. We report the performance on different approaches on aspect summarization in Table~\ref{tab:asp-results} on {\OPO} and {\SPACE}.
% \footnote{{\AMZN} dataset does not contain any aspect-specific annotations, or corresponding summaries.} 
We observe that {\X} achieves the state-of-the-art performance for all metrics on the {\OPO} dataset. On {\SPACE} dataset, it achieves comparable scores to other extractive approaches.

\noindent \textbf{Human Evaluation}. We perform a human evaluation to compare the summaries from {\X} with the state-of-the-art extractive summarization systems SemAE and QT. General summaries were judged based on the following criteria: \textit{informativeness}, \textit{coherence}, and \textit{redundancy}. We present human evaluators with summaries in a pairwise fashion and ask them to select which one was better/worse/similar according to the criteria. The final scores for each system reported in Table~\ref{tab:human-eval-general} were computed using Best-Worst Scaling~\cite{louviere2015best}. We observe that {\X} outperforms the baselines in coherence and redundancy. {\X} performs slightly worse than QT in informativeness. This is expected as {\X} greedily select sentences (that are often similar), while QT performs sampling leading to more coherent  summaries (compromising on informativeness).

\begin{table}[t!]
    % \small
	\centering
    \resizebox{0.35\textwidth}{!}{
	\begin{tabular}{l c  c c} 
		\toprule[1pt]
		 Aspect  & Exclusive & Partial & None \\ 			
		\midrule[0.5pt]
        {SemAE} & 22.1 & {43.8} & 34.1\\
		QT &  22.2 & 41.9 & 35.9\\
		{\X} & ~~\textbf{46.4}* & \textbf{45.6} & ~~~\textbf{8.0}*\\
		\bottomrule[1pt]
	\end{tabular}
	}
    \vspace{-6pt}
	\caption{Human evaluation results of aspect summarization for {\OPO} dataset. {\X} generates more aspect-specific summaries compared to baselines.}
	\label{tab:human-eval-aspect}
	\vspace{-10pt}
\end{table}

For aspect summaries, we ask annotators to judge whether a summary discusses a specific aspect \textit{exclusively}, \textit{partially}, or \textit{does not mention} it at all. 
In  Table~\ref{tab:human-eval-aspect}, we report the human evaluation results for aspect summaries on {\OPO} dataset. {We observe that {\X} generates summaries that are significantly more aspect-specific compared to baselines.} We provide further details about human evaluation in Appendix~\ref{sec:human-eval}.

\section{Analysis}

\noindent \textbf{Thawed Encoder}.\label{sec:frozen} In this experiment, we compare the performance of {\X} when the encoder is allowed to be fine-tuned with the original setup, where the encoder weights are frozen. In Table~\ref{tab:thawed}, we observe that there is a significant drop in performance when the encoder is fine-tuned. We hypothesize that this happens because the model overfits shallow word-level semantics, and is unable to capture more abstract semantics. This showcases the utility of pre-trained representations that helps {\X} perform well in an unsupervised setting.

Next, we investigate the efficacy of the representation learning and sentence selection modules by replacing each of them with a competitive variant.

\begin{table}[t!]
    \small
    \centering
	\resizebox{0.45\textwidth}{!}{
		\begin{tabular}{l c c c } 
			\toprule[1pt]
			 Dataset & R1 & R2 & RL\\
			\midrule[1pt]
			\multirow{1}{*}{\OPO}
			&  35.7 \footnotesize{(\textcolor{red}{$\downarrow$5.9})} & 13.9 (\textcolor{red}{$\downarrow$6.9}) & 21.1 (\textcolor{red}{$\downarrow$4.1})\\
			\multirow{1}{*}{\AMZN}
			&  32.2 (\textcolor{red}{$\downarrow$1.6})   & ~~6.2 (\textcolor{red}{$\downarrow$1.0}) & 17.3 (\textcolor{red}{$\downarrow$1.5})\\
			\multirow{1}{*}{\SPACE}
			&  33.5 (\textcolor{red}{$\downarrow$8.9}) & ~~6.9 (\textcolor{red}{$\downarrow$5.5}) & 19.5 (\textcolor{red}{$\downarrow$5.3})\\
			\bottomrule[1pt]
	\end{tabular}
}
	\vspace{-3pt}
\caption{Evaluation results when {\X}'s encoder is fine-tuned during training. We observe a significant drop in performance when the encoder is fine-tuned.}
\label{tab:thawed}
\vspace{-10pt}
\end{table}

\begin{table}[h!]
    \small
    \centering
	\resizebox{0.48\textwidth}{!}{
	\begin{tabular}{l c c c } 
			\toprule[1pt]
			 Dataset & R1 & R2 & RL\\
			\midrule[1pt]
			{\OPO} & ~~28.1 (\textcolor{red}{\footnotesize $\downarrow$13.5}) & ~~~~6.2 (\textcolor{red}{\footnotesize $\downarrow$14.6}) & ~15.8  (\textcolor{red}{\footnotesize $\downarrow$9.4})\\
            {\AMZN} & 32.6 (\textcolor{red}{\footnotesize $\downarrow$1.2})  &  ~~{6.1} (\textcolor{red}{\footnotesize $\downarrow$1.1})  & ~17.9 (\textcolor{red}{\footnotesize $\downarrow$0.8})  \\
            {\SPACE} & 41.4 (\textcolor{red}{\footnotesize $\downarrow$1.0}) & 11.5 (\textcolor{red}{\footnotesize $\downarrow$0.9}) & ~24.0  (\textcolor{red}{\footnotesize $\downarrow$0.8}) \\
			\bottomrule[1pt]
	\end{tabular}
	}
	\vspace{-6pt}
\caption{Evaluation results of {\X} with a modified score $I(s) = -\| \mathbf{x}_s - \mu_e\|^2_2$. We observe a significant drop in performance across all three datasets.}
\label{tab:euclidean}
\vspace{-5pt}
\end{table}

\noindent \textbf{Euclidean-based Importance Score}. 
We investigate the utility of geodesic-based importance scoring over Euclidean-based scoring.
In this experiment, instead of $I(s)$ (defined in Equation~\ref{eqn:rel})
we compute the importance score of a sentence, $s$, as the Euclidean distance from the mean representation, $\mu_e$ ($I(s) = -\| \mathbf{x}_s - \mu_e\|^2_2$). 
We report the results of this setup in Table~\ref{tab:euclidean} (relative performance to {\X} is shown in brackets). 
We observe that performing sentence selection using Euclidean distance results in a significant drop in performance across all datasets. We believe that leveraging the $k$NN graph provides us with a better approximation of the underlying representation manifold, which results in better summarization performance.

\begin{table}[t!]
    \small
    \centering
	\resizebox{0.49\textwidth}{!}{
		\begin{tabular}{l c c c c } 
			\toprule[1pt]
			 Dataset & Model & R1 & R2 & RL\\
			\midrule[1pt]

            \multirow{2}{*}{\OPO} 
            & RBT & 35.1 (\textcolor{red}{\footnotesize $\downarrow$6.5}) & 13.0 (\textcolor{red}{\footnotesize $\downarrow$7.8}) & 21.2 (\textcolor{red}{\footnotesize $\downarrow$4.0})\\
			& SCS & 33.5 (\textcolor{red}{\footnotesize $\downarrow$8.1}) & ~~6.9 (\textcolor{red}{\footnotesize $\downarrow$13.9}) & ~19.5 (\textcolor{red}{\footnotesize $\downarrow$5.7}) \Bstrut\\
			\hline
            \multirow{2}{*}{\AMZN} 
            & RBT & 29.4 (\textcolor{red}{\footnotesize $\downarrow$4.4})  & 4.7 (\textcolor{red}{\footnotesize $\downarrow$2.4})  & 15.3 (\textcolor{red}{\footnotesize $\downarrow$3.5})  \Tstrut\\
            & SCS & 31.0 (\textcolor{red}{\footnotesize $\downarrow$2.8}) & 5.2 (\textcolor{red}{\footnotesize $\downarrow$1.9}) & 16.4 (\textcolor{red}{\footnotesize $\downarrow$2.4}) \Bstrut\\
            \hline
            \multirow{2}{*}{\SPACE} 
            & RBT & ~26.8 (\textcolor{red}{\footnotesize $\downarrow$15.6}) & 3.7 (\textcolor{red}{\footnotesize $\downarrow$8.7}) & ~15.4  (\textcolor{red}{\footnotesize $\downarrow$9.4}) \Tstrut\\
            & SCS & ~30.1 (\textcolor{red}{\footnotesize $\downarrow$12.3}) & 4.8 (\textcolor{red}{\footnotesize $\downarrow$7.6}) & 17.3 (\textcolor{red}{\footnotesize $\downarrow$7.5})\\
			\bottomrule[1pt]
	\end{tabular}}

	\vspace{-5pt}
\caption{ Evaluation results of {\X} using RoBERTa (RBT) and SimCSE's  (SCS) representations. We observe that opinion summarization using topical representations from {\X} outperforms distributed representations across all datasets.}
\label{tab:distributed}
%\vspace{-10pt}
\end{table}

\noindent \textbf{Distributed vs. Topical Representations}. In this experiment, we investigate the relative efficacy of topical representations compared to distributed representations. We retrieve distributed sentence representations from RoBERTa~\cite{roberta} (\texttt{[CLS]} token feature) and SimCSE~\cite{simcse} model. Then, we use these representations in our sentence selection algorithm (Section~\ref{subsec:ss}) to compose the summary. In Table~\ref{tab:distributed},
 we observe that topical representations (obtained from {\X}) outperform distributed representations by a significant margin across all setups.
 This shows the utility of topical representations over distributed representations for unsupervised summarization.

\begin{table}[t!]
    \centering
    \resizebox{0.36\textwidth}{!}{
    \begin{tabular}{l c c}
    \toprule
        Perplexity (PPL) $\downarrow$ & {\SPACE}  & {\AMZN}  \\
    \midrule
        QT & 33.46 & 63.70\\
        SemAE & 15.95 & 55.46\\
        {\X} & \textbf{14.95} & \textbf{45.55}\\
    \bottomrule
    \end{tabular}
    }
    \caption{Perplexity of the summaries generated by different extractive summarization systems. We observe that {\X} achieves the best perplexity scores, indicating more coherent summaries.}
    \label{tab:coherence}
    \vspace{-10pt}
\end{table}

\noindent \textbf{Summary Coherence}. In this experiment, we evaluate the coherence of the generated extractive summaries using automatic measures. Specifically, we measure the perplexity scores (from HuggingFace~\cite{hf} Evaluate API) using the GPT-Neo model~\cite{gpt-neo}. The perplexity scores are indicative of the coherence of the generated text. In Table~\ref{tab:coherence}, we report the perplexity scores on {\SPACE} and {\AMZN} datasets for extractive systems QT, SemAE, and {\X}. We observe that {\X} achieves the best perplexity scores showcasing that it is able to generate superior-quality summaries in terms of coherence. We believe that the greedy aggregation of sentences in {\X} often results in the selection of semantically similar sentences thereby leading to more coherent summaries with fewer context switches.

\noindent \textbf{Cluster Interpretation}. In this experiment, we investigate whether different parts of the representation space capture distinct semantics. We partition the space by performing agglomerative clustering with Ward's linkage~\cite{ward1963hierarchical}  on the representation set for a particular entity. 
In Table~\ref{tab:semantics}, we report example sentences within different clusters. We observe that sentences belonging to the same cluster share a common theme. The underlying semantics of a cluster can vary from being coarse, like the presence of the phrase `Calistoga', to more nuanced concepts like pillows \& beds in the room, flowers in the hotel's garden, etc.

\noindent \textbf{Generated Summaries}. In Table~\ref{tab:summaries}, we report the summaries generated by {\X}, and other comparable extractive summarization systems like SemAE and QT. We observe that {\X} is able to generate a comprehensive summary that reflects the main considerations mentioned in the human summary. Compared to SemAE, we see more specific adjectival descriptions; SemAE indicates that many of the hotel characteristics are simply `great'. Compared to QT, we see a review that seems to more accurately reflect the human-written summary.

We perform additional ablations experiments to investigate the domain transfer capabilities, sparsity of representations, among others in Appendix~\ref{sec:analysis}.

\begin{table}[t!]
    \small
    \centering
	
\begin{tabular}{p{0.1\textwidth} | p{0.30\textwidth}} 
			\toprule[1pt]
			 \textbf{Theme} & \multicolumn{1}{|c}{\textbf{Sentences}}  \TBstrut\\
			\midrule[1pt]
     {\hspace{12cm} Flowers} &			
     \tabitem The gardens are lovely with wide varieties of {\sethlcolor{lightgreen}\hl{flowering plants}} and shrubs, koi ponds and hummingbird feeders.
     
     \tabitem \sethlcolor{lightgreen}\hl{Pots of tulips and daffodils} in full bloom; other plantings well cared for; pathways clean and swept.
\\
     \midrule[1pt]
      \hspace{3.2cm}Location `Calistoga' & 
      \tabitem {\sethlcolor{lightgreen}\hl{Calistoga}} is a beautiful historic town with good restaurants and beautiful old houses --a fun place to walk.
     
    \tabitem The Roman Spa and {\sethlcolor{lightgreen}\hl{Calistoga}} is our favorite spot in the Wine Country.
         \\
        \midrule[1pt]
        \hspace{1.5cm}Pillows \& Beds & \tabitem The rooms were in great shape, very clean, comfortable {\sethlcolor{lightgreen}\hl{beds}} with lots of {\sethlcolor{lightgreen}\hl{pillows}}.
        
        \tabitem The {\sethlcolor{lightgreen}\hl{pillows}} and {\sethlcolor{lightgreen}\hl{bed}} coverings were of very good quality There was also a mini-refrigerator and coffeemaker.
  \\
   \midrule[1pt]
        \hspace{1.5cm}Phrase `every year' & \tabitem We return {\sethlcolor{lightgreen}\hl{every year}} to the Roman Space after the holidays and brought Seattle friends this January.
        
        \tabitem {\sethlcolor{lightgreen}\hl{Every year}} for the past 15 years we have met at the Roman Spa ... 
  \\
			\bottomrule[1pt]
	\end{tabular}
	\vspace{-5pt}
\caption{ Sentences within a cluster produced from agglomerate clustering. Sentences in a row belong to the same cluster. We highlight the dominant theme of a cluster in green. }
\label{tab:semantics}
\vspace{-15pt}
\end{table}

\section{Related Work}

\begin{table*}[t!]
    	\centering
	\resizebox{\textwidth}{!}{
    	\footnotesize
    	\begin{tabular}{p{0.23\textwidth} | p{0.21\textwidth} | p{0.22\textwidth} | p{0.2\textwidth}} %{ l|l|l|l} 
		\toprule[1pt]
		\multicolumn{1}{c|}{\textbf{Human}} & \multicolumn{1}{c|}{\textbf{\X}} & \multicolumn{1}{c|}{\textbf{SemAE}} & \multicolumn{1}{c}{\textbf{QT}} \TBstrut\\ 			
		\midrule[1pt]
		& & & \\[-0.5em]
		All staff members were friendly, accommodating, and helpful. The hotel and room were very clean. The room had modern charm and was		nicely remodeled. The beds are extremely comfortable. The rooms are	quite with wonderful beach views. The food at Hash, the restaurant in lobby, was fabulous. The location is great, very close to the beach. It’s a	longish walk to Santa Monica. The price is very affordable. 
		&  {\sethlcolor{lightgreen}\hl{Overall we had a nice stay at the hotel.}} Our room was very clean and comfortable. The atmosphere is stylish and the service was great. We ate breakfast at the hotel and it was great. I appreciate the location and the security in the hotel. The food and service at the restaurant was awesome. The Hotel is classy and has a rooftop bar. The restaurant is cozy but they have good healthy food. Great hotel.
		& The staff is {\sethlcolor{LightRed}\hl{great}}.	The Hotel Erwin is a {\sethlcolor{LightRed}\hl{great}} place to stay.	The staff were friendly and helpful.	The location is perfect.	We ate breakfast at the hotel and it was {\sethlcolor{LightRed}\hl{great}}.	The hotel itself is in a {\sethlcolor{LightRed}\hl{great}} location.	{{The service was wonderful.	It was {\sethlcolor{LightRed}\hl{great}}.	The rooms are {\sethlcolor{LightRed}\hl{great}}.	The rooftop bar HIGH was the icing on the cake.}}	The food and service at the restaurant was awesome.	The service was excellent.
		& Great hotel. We liked our room with an ocean view. {{The staff were friendly and helpful. There was no balcony. The location is perfect. Our room was very quiet.}} I would definitely stay here again. You’re one block from the beach. So it must be good! {\sethlcolor{LightRed}\hl{Filthy hallways.		Unvacuumed room.}} {\sethlcolor{LightRed}\hl{Pricy,		but well worth it}}.
		\\
		\bottomrule[1pt]
\end{tabular}

	}
 \vspace{-5pt}
	\caption{ Human-written  and generated summaries from {\X}, SemAE, and QT. For a fair comparison, we present the summary for the instance reported in previous works. {\X} generates a comprehensive review with a relatively logical ordering that starts with a clear topic sentence and then proceeds to details. Compared to SemAE, we see more descriptive sentences selected. Compared to QT, we see a summary that more closely matches the human-written summary.}
    \label{tab:summaries}
    \vspace{-10pt}
\end{table*}
Most work on opinion summarization focuses on generating summaries in an unsupervised setup due to the scarcity of labeled data. These works are broadly classified into two categories based on the type of summaries being generated: \textit{abstractive}~\cite{ganesan2010opinosis, carenini2006multi, di2014hybrid} or \textit{extractive}~\cite{erkan2004lexrank, nenkova2005impact, zhao2022read, li2023aspect}. Abstractive systems, in an unsupervised  setup~\cite{chu2019meansum, bravzinskas2019unsupervised, iso-etal-2021-convex-aggregation, wang-wan-2021-transsum, amplayo-etal-2021-aspect} train an encoder-decoder setup using a self-supervised objective and generate the summary by leveraging the aggregate opinion representation.
On the other hand, extractive opinion systems~\cite{kim2011comprehensive}, select sentences using an importance score that quantifies their salience. Salience has been computed using frequency-based approaches~\cite{nenkova2005impact}, distance from mean~\cite{radev2004centroid}, or graph-based techniques~\cite{erkan2004lexrank}. Few approaches focus on aspect specificity and sentiment polarity for sentence selection~\cite{angelidis2018summarizing, zhao2020weakly}.

Our work is most similar to extractive summarization systems SemAE~\cite{semae}, and QT~\cite{qt}. Similar to these systems, {\GS} has two components: a representation learning system, and a sentence selection routine. However, unlike these approaches, we leverage pre-trained models to learn topical representations over a latent dictionary and propose a sentence selection mechanism using approximate geodesics to perform summarization.

Approaches in our work resemble prior works in deep clustering, which considers a similar combination of unsupervised representation learning and sparse structures~\cite{yang2016joint,jiang2016variational, law2017deep, caron2020unsupervised,zhao2020unsupervised}. In a similar fashion, dictionary learning-like approaches have been combined with deep networks \cite{liang2020anchor,zheng2021deep} for various tasks.

%\vspace{-1mm}
\section{Conclusion}
%\vspace{-1mm}

We present {\GS}, a novel framework for extractive opinion summarization. % by learning topical sentence representations in an unsupervised manner. 
{\X} uses a representation learning model to convert distributed representations from a pre-trained model into topical text representations. {{\X} uses these representations to compute the importance of a sentence using approximate geodesics. We show that {\X} achieves strong performance on several opinion summarization datasets.}
However, there are a lot of open questions about the inductive biases of representation learning that are needed for unsupervised summarization. In this work, we show the efficacy of topical representations. However, are there better approaches to capturing language semantics that help us quantify the importance of an opinion?
Our analysis shows that representations from {\X} span the high-dimensional space in a manner that different parts of it capture distinct semantics. 
This opens up the possibility of leveraging the representation geometry to capture different forms of semantics.  Future work can explore ways to leverage topical representations from {\X} for tasks where there is a scarcity of labeled data.

\section{Acknowledgement}
The authors are thankful to Anneliese Brei, Haoyuan Li, Anvesh Rao Vijjini, and Chao Zhao
 for helpful feedback on an earlier version of this paper.  The work of Somnath Basu Roy Chowdhury and Snigdha Chaturvedi was supported in part by the National Science Foundation under award DRL-2112635.

\section{Limitations}
We propose {\X}, a novel system that learns topical representations of text and uses them to compute the importance of opinion reviews for extractive summarization.
One of the limitations of {\X} is {that} it requires pre-training of the representation learning module using reviews sentences from a similar domain. For this, {\X} requires access to a large collection of review data from the target domain, thereby limiting its applicability in zero-shot or few-shot setups. This can be alleviated by future research on developing foundational models that learn topical representations on large-scale datasets and generalize across different opinion summarization domains.

\section*{Ethical Considerations}
We do not foresee any ethical issues from the technology introduced in this paper. However, we would like to mention certain limitations of extractive summarization systems in general. As extractive systems select review sentences from the input, it can produce undesirable output when the input reviews have foul or offensive language. Therefore, it is important to remove foul language from the input in order to ensure the end user is not affected.  In general, we use public datasets and do not annotate any data manually. All datasets used in this paper have customer reviews in the English language. Human evaluations for summarization were performed on Amazon Mechanical Turks (AMT) platform. Human judges were based in the United States. Human judges were compensated at a rate of at least \$15 USD per hour.

\bibliography{anthology,custom}

\clearpage
\appendix
\section{Appendix}

\subsection{Human Evaluation}
\label{sec:human-eval}
We perform the human evaluation on the Amazon Mechanical Turk (AMT) platform. We designed the payment rate per Human Intelligence Task (HIT) in a manner to ensure that judges were compensated at a rate of at least \$15 USD per hour. In all tasks, each HIT was evaluated by three human judges.

For general summarization, we performed a pairwise evaluation of two summarization systems. Specifically, we were given two system summaries the human judges were asked to judge each pair  as better, worse, or similar. We asked the judges to evaluate the pair based on the following criteria -- \textit{informativeness}, \textit{redundancy}, and \textit{coherence}, in independent tasks. For informativeness, we also provide the judges with a human-written summary. The judges annotate a summary as more informative only if the information is consistent with the human-written summaries. The reported scores (-100 to +100) were computed using Best-worst scaling~\cite{louviere2015best}. { For a fair comparison, we consider the version of SemAE that does not use additional aspect-related information.}

For aspect summarization, we provide human judges with a system-generated aspect summary and the corresponding aspect. Judges were asked to annotate whether the system summary discusses the mentioned aspect \textit{exclusively}, \textit{partially}, or \textit{does not mention} the aspect at all.

\subsection{Analysis}
\label{sec:analysis}
\noindent \textbf{Dictionary Size Ablation}. In this experiment, we vary the number of elements in each dictionary ($m$) and observe the summarization performance on {\SPACE} dataset. We conduct these experiments on the {\SPACE} dataset. In Table~\ref{tab:dict_size}, we observe {\X} achieves comparable performance with significantly smaller dictionary sizes. In fact, for the smallest dictionary sizes {\X} achieves the best ROUGE-1 and ROUGE-L scores.

\begin{table}[t!]
    \small
    \centering
	\begin{tabular}{l c c c } 
    \toprule[1pt]

     $m$ & R1 & R2 & RL\\
    \midrule[1pt]
    512 & \textbf{43.36} & 11.53 & {24.10}\\
    1024 & 42.76 & \textbf{12.66} & 24.28\\
    4096 & {42.77} & {11.47} & {24.11}\\
    8192 & 42.36 & 12.44 & \textbf{24.80}\\
    16384 & 41.24 & 10.92 & 23.92\\
    \bottomrule[1pt]
\end{tabular}

\caption{ Evaluation results with a varying number of dictionary elements on {\SPACE} dataset. We observe that there is only a small drop in performance of {\X}, when the dictionary sizes are reduced.
}
\label{tab:dict_size}
\end{table}

\begin{table*}[t!]
    \centering
    \resizebox{0.93\textwidth}{!}{
		\begin{tabular}{ l c c c | c c c | c c c } 
			\toprule[1pt]
            \multirow{2}{*}{Method} & \multicolumn{3}{c|}{{\OPO}} & \multicolumn{3}{c|}{{\AMZN}} & \multicolumn{3}{c}{{\SPACE}}\\
			& R1 & R2 & RL & R1 & R2 & RL & R1 & R2 & RL \\ 	
			\midrule[1pt]		
			NMF~\cite{nmf} & 32.85 & 10.44 & 18.96 & 30.33 & 5.07 & 16.10 & 34.88 & 6.14 & 18.87\Tstrut\\
	    	LDA~\cite{lda}\hspace*{0.15cm} & 32.70 & 10.85& 19.60 & 31.31 & 5.27 &16.51  & 26.57 & 3.46 & 14.81 \\
			LSA~\cite{lsa} & 32.41 & 10.33 & 19.66 & 31.71 & 6.11 & 17.79 & 31.64 & 5.57 & 17.72\\
			HDP~\cite{hdp} & 34.60 & 11.29 & 19.39 & 30.60 & 4.91 & 16.20 & 29.77 & 4.44 &  16.49    \\
			NTM\textsubscript{BERT}~\cite{ntm} & 33.00 & 11.01 & 19.01 & 31.62 & 5.29 & 16.54 & 26.12 & 2.74 &  15.29  \Bstrut\\
			\midrule[1pt]
	 {\GS} (\X) & \textbf{41.55}  & \textbf{20.77} & \textbf{25.19} & \textbf{33.75}  & \textbf{7.15} &  \textbf{18.79} & \textbf{42.36} & \textbf{12.44} & \textbf{24.80} \TBstrut\\
			\bottomrule[1pt]
	\end{tabular}
}

    \caption{Comparison of {\GS}'s performance with other unsupervised topic modeling techniques on general summarization. In this experiment, we modify the representation learning module of {\GS} while keeping the sentence selection approach same. We observe that {\GS}'s topic modeling approach achieves the best performance across all datasets.}
    \label{tab:topical}
\end{table*}

\begin{table}[t!]
    \centering
	\resizebox{0.45\textwidth}{!}{
		\begin{tabular}{l c c c } 
			\toprule[1pt]
			 Train$\rightarrow$Predict & R1 & R2 & RL\\
			\midrule[1pt]
			{\SPACE}$\rightarrow${\OPO} & 38.94 & 16.80 & 22.60 \\
			{\AMZN}$\rightarrow${\OPO} & 40.14 & 18.97 & 24.91 \\
			{C4}$\rightarrow${\OPO} & 37.96 & 15.97 & 21.93 \\
			\rowcolor{Gray}
			{\OPO}$\rightarrow${\OPO} & \textbf{41.55}  & \textbf{20.77} & \textbf{25.19}  \\
			\midrule[0.5pt]
			{\SPACE}$\rightarrow${\AMZN} & 32.29 & 6.36 & 17.22 \\
			{\OPO}$\rightarrow${\AMZN} & 33.57  & 6.46  & 17.86  \\
			{C4}$\rightarrow${\AMZN} & 32.03  & 6.35 & 17.04 \\
			\rowcolor{Gray}
			{\AMZN}$\rightarrow${\AMZN} & \textbf{33.75}  & \textbf{7.15} &  \textbf{18.79}   \\
			\midrule[0.5pt]
			{\OPO}$\rightarrow${\SPACE} & 27.85  & 4.93  & 16.03  \\
			{\AMZN}$\rightarrow${\SPACE} & 25.14 & 2.95 & 14.98  \\
			{C4}$\rightarrow${\SPACE} & 41.81 & 11.61 & 24.28  \\
			\rowcolor{Gray}
			{\SPACE}$\rightarrow${\SPACE} & \bf {42.36} & \bf 12.44 & \textbf{24.80}  \\
			\bottomrule[1pt]
	\end{tabular}
}

\caption{ Evaluation results when the representation learning system is trained on a different dataset. 
In-domain performance is highlighted in \colorbox{Gray}{gray}. {\X} shows decent domain transfer performance for {\OPO} and {\AMZN} datasets. }
\label{tab:transfer}
\end{table}

\noindent \textbf{Sparsity}. 

\begin{figure}[h!]
    \centering
    \includegraphics[width=0.4\textwidth, keepaspectratio]{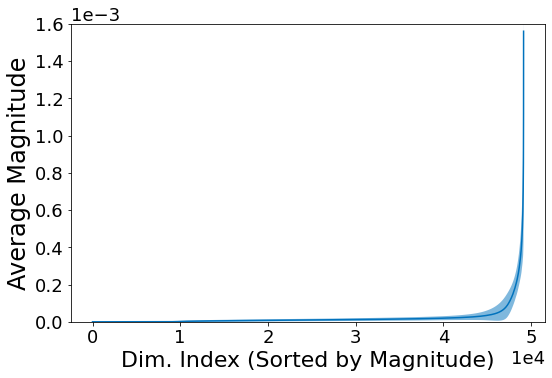}
    
    \caption{Plot depicting the sparsity of sentence representations retrieved from {\X}. We sort, individually for each sentence, the dimensions from the smallest to the largest magnitude and report the mean magnitude for each sorted position (and two standard deviations). Most sentences seem to have only a few large magnitude dimensions and many close to zero. }
    \label{fig:sparsity}
\end{figure}

We examine the sparsity of sentence representations from {\X}. For each sentence representation, we sort the dimensions by magnitude, from smallest to largest. This enables us to compare magnitudes across sentences for a specific sorted rank position. We then plot the mean magnitude (and two standard deviations) for each sorted rank position, as illustrated in Figure~\ref{fig:sparsity}. Our observations indicate that most sentences possess only a few dimensions with high magnitude, while the remaining dimensions have magnitudes of zero or close to zero.

\noindent \textbf{Domain Transfer capability}.
In this experiment, we investigate the domain transfer capability of {\X}. Specifically, we evaluate how {\X} trained on one dataset, performs on others. We also evaluate {\X} when it is trained on C4 dataset~\cite{raffel2020exploring}. In Table~\ref{tab:transfer}, we report the results of this experiment. We observe that when training on the non-domain specific C4 corpus, performance is nearly that of in-domain training. The largest degradation of performance occurs when training on  {\OPO} or {\AMZN} and evaluating on {\SPACE}. We hypothesize that this happens due to a domain shift, where both {\AMZN} and {\OPO} are product review datasets, while {\SPACE} has reviews for hotel entities. When evaluated on {\OPO} or {\AMZN},  we observe that {\X} is generalizing well, and out-of-domain performance is not much worse than in-domain performance (highlighted in \colorbox{Gray}{gray}).

\noindent\textbf{Unsupervised Topic Modeling Ablations}
In this setup, we experiment with different unsupervised topic modeling approaches -- latent Dirichlet allocation (LDA)~\cite{lda}, linear semantic analysis (LSA)~\cite{lsa}, non-negative matrix factorization (NMF)~\cite{nmf}, hierarchical Dirichlet process (HDP)~\cite{hdp}, and neural topic model (NTM) using contextual embeddings~\cite{ntm}. Most of these approaches focus on factorizing sentence representations into topical representations over a set of learned topics. We set the sentence representation dimension $d=100$ for all approaches. Specifically, we replace the representation learning module from {\X} while keeping the sentence selection algorithm the same. In Table~\ref{tab:topical}, we report the performance on general summarization of different methods. We observe that most of the other topical approaches perform significantly worse than {\X}. These approaches use significantly fewer parameters compared to a Transformer decoder used in {\X}. We believe that leveraging more parameters helps the unsupervised model to capture latent semantics better leading to better summarization performance.

\end{document}